\documentclass{article}

\PassOptionsToPackage{numbers, compress}{natbib}

     \usepackage[final]{neurips_2019}

\usepackage[utf8]{inputenc} 
\usepackage[T1]{fontenc}    
\usepackage{hyperref}       
\usepackage{url}            
\usepackage{booktabs}       
\usepackage{amsfonts}       
\usepackage{nicefrac}       
\usepackage{microtype}      
\usepackage{graphicx}
\usepackage{subfigure}
\usepackage{algorithm}
\usepackage{algpseudocode}

\usepackage{paralist}
\usepackage[textsize=scriptsize,textwidth=3.2cm]{todonotes}

\title{MARLeME: A Multi-Agent Reinforcement Learning Model Extraction Library}

\author{%
  Dmitry Kazhdan \\
  Dept. of Computer Science and Technology \\
  University of Cambridge\\
  \texttt{dk525@cam.ac.uk} \\
   \And
   Zohreh Shams \\
   Dept. of Computer Science and Technology \\
   University of Cambridge \\
   \texttt{zohreh.shams@cl.cam.ac.uk} \\
   \AND
   Pietro Li\`{o} \\
   Dept. of Computer Science and Technology \\
   University of Cambridge \\
   \texttt{pietro.lio@cl.cam.ac.uk}
}

\begin{document}

\maketitle

\begin{abstract}

Multi-Agent Reinforcement Learning (MARL) encompasses a powerful class of methodologies that have been applied in a wide range of fields. An effective way to further empower these methodologies is to develop libraries and tools that could expand their interpretability and explainability. In this work, we introduce MARLeME: a MARL model extraction library, designed to improve explainability of MARL systems by approximating them with symbolic models. Symbolic models offer a high degree of interpretability, well-defined properties, and verifiable behaviour. Consequently, they can be used to inspect and better understand the underlying MARL system and corresponding MARL agents, as well as to replace all/some of the agents that are particularly safety and security critical.

\end{abstract}

\section{Introduction}

Multi-Agent Reinforcement Learning (MARL) has achieved groundbreaking results in a wide range of fields, and is currently a highly active research area of Machine Learning (ML) \citep{Pouyanfar:2018:SDL:3271482.3234150, arulkumaran2017brief}. MARL deals with teams of agents that learn how to act optimally in stochastic environments through trial-and-error, and has been successfully applied to tasks requiring cooperative/competitive multi-agent behaviour, such as large-scale fleet management \citep{lin2018efficient}, swarm systems \citep{huttenrauch2017guided}, and task allocation \citep{noureddine2017multi}.

Unfortunately, the vast majority of existing MARL approaches represent decision-making policies using very complex models, such as Deep Neural Networks (DNNs) \citep{li2017deep}, making it impossible to directly understand an agent's action-selection strategy. A lack of interpretability of such systems leads to a lack of confidence in the correctness of their behaviour, which is crucial in safety-critical applications, such as self-driving cars or healthcare. Furthermore, this lack of interpretability implies that optimal strategies learned by the MARL agents cannot be used to improve our understanding of the corresponding domain. Approaches based on symbolic reasoning, on the other hand, offer interpretable, verifiable models with well-defined properties. As a result, there has recently been increasing interest in approaches capable of combining symbolic reasoning with ML \cite{Domingos:2019:ULS:3342113.3241978, evans2018learning, garcez2019neural, rocktaschel2017end}.

 One technique that allows reaping the benefits of both ML-based and symbolic approaches is \textit{model extraction} \citep{DBLP:journals/corr/BastaniKB17a}. Model extraction refers to approaches that approximate a complex model (e.g. a DNN) with a simpler, interpretable one (e.g. a rule-based model), facilitating understanding of the complex model. Intuitively, statistical properties of the complex model should be reflected in the extracted model, provided approximation quality of the extracted model (referred to as \textit{fidelity}) is high enough.
 

This paper introduces MARLeME: a (M)ulti-(A)gent (R)einforcement (Le)arning (M)odel (E)xtraction library, designed to improve interpretability of MARL systems using model extraction. MARLeME is an open-source~\footnote{https://github.com/dmitrykazhdan/MARLeME}, easy-to-use, plug-and-play library, that can be seamlessly integrated with a wide range of existing MARL tasks, in order to improve explainability of the underlying MARL system. To the best of our knowledge, this is the first open-source MARL library focusing on interpretable model extraction from MARL systems.

The rest of this paper is structured as follows: Section \ref{rel_work} reviews work related to RL model extraction. Section \ref{marleme_sec} gives an overview of the MARLeME library.  
Section \ref{eval} and Section \ref{results} present an evaluation of MARLeME using two RL benchmarks (Mountain Car and RoboCup Takeaway). Finally, Section \ref{conclusion} gives some concluding remarks. 



\section{Related Work} \label{rel_work}

A range of recent work has focused on Single-Agent Reinforcement Learning (SARL) interpretable model extraction. Work in \citep{verma2018programmatically} generates interpretable and verifiable agent policies represented using a high-level, domain-specific programming language. Work in \citep{hein2018interpretable} uses model-based batch RL and genetic programming to learn policy equations. Finally, work in \citep{brown2018interpretable} uses boosted regression tree techniques to compute human-interpretable policies from a RL system. Unlike MARLeME, which is designed to work with teams of RL agents, all of the above approaches were designed for and evaluated using the single RL agent case only.

Work in \citep{tacchetti2018relational} presents an approach to modelling teams of agents using Graph Neural Networks (GNNs), by training GNNs to predict agent actions, and then using their intermediate representations (e.g. the magnitudes of the edge attributes) to study the social dynamics of the MARL system (e.g. the degree of influence agents have on each other). This approach interprets the original GNN model directly, through a post-hoc analysis of its constituent parts, whereas MARLeME relies on approximating original models with simpler, interpretable ones. Crucially, GNN intepretation is still a relatively unexplored field, lacking well-established principles and best-practices. MARLeME on the other hand, makes use of well-studied and well-understood symbolic reasoning models when analysing MARL systems.  


Our work is also partially related to imitation learning \citep{hussein2017imitation, kober2010imitation}. Similarly to MARLeME, imitation learning approaches derive action-selection policies from input \textit{trajectories} (an agent trajectory consists of a sequence of states and corresponding selected actions collected over a set of episodes). In particular, work described in \citep{le2017coordinated} learns a coordination model, along with the individual policies, from trajectories of multiple coordinating agents, using imitation learning and unsupervised structure learning. Crucially, however, imitation learning, including the approach used in \citep{le2017coordinated}, focuses on the quality of the policy (i.e. its performance on the corresponding task), without paying attention to its interpretability. Achieving the latter while preserving performance is a unique focus of MARLeME.

\section{MARLeME} \label{marleme_sec}

\begin{figure}
  \centering
  \includegraphics[scale=0.35]{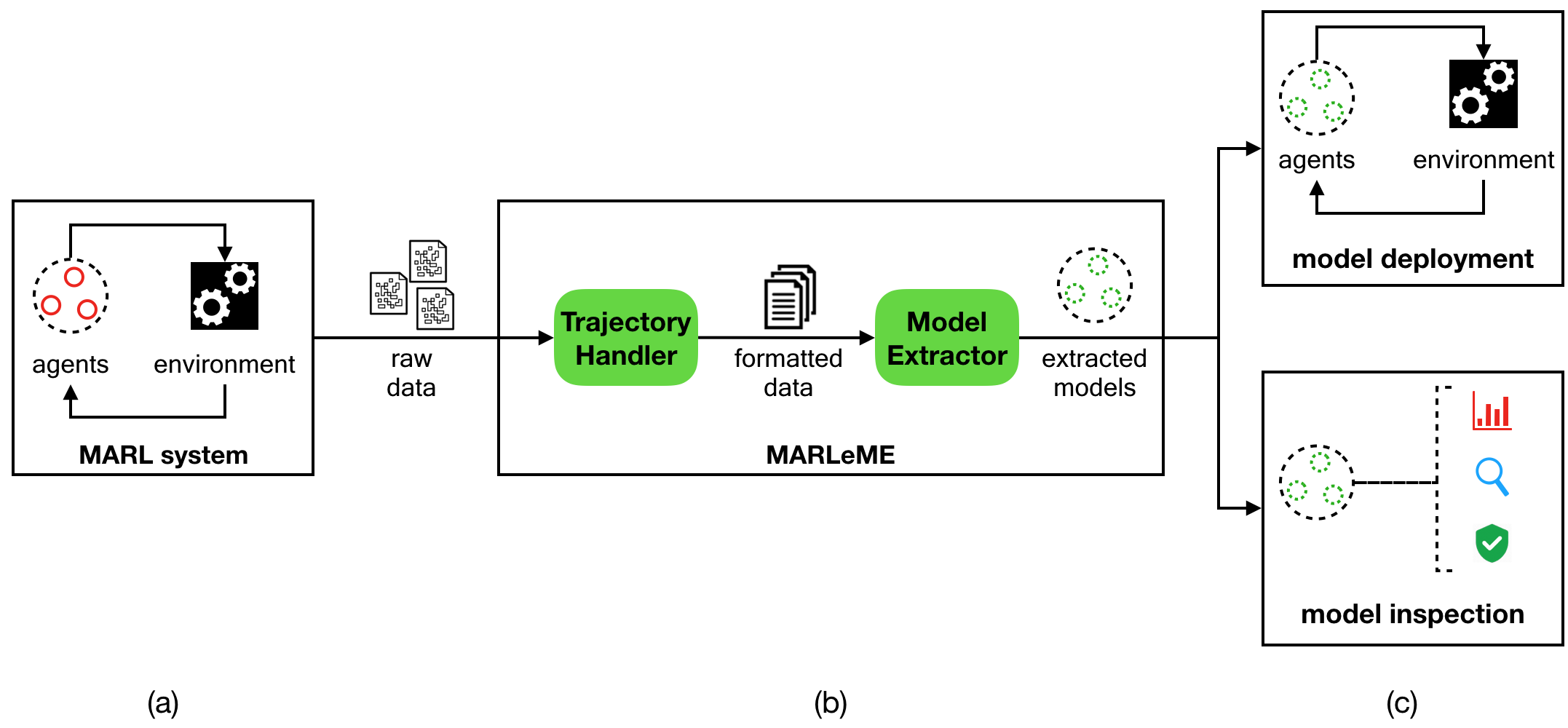} 
\caption{(a) A MARL system consists of a set of RL agents (red circles) interacting with each other and their environment. (b)  MARLeME uses agent trajectory data obtained from the MARL system (raw data), in order to extract a set of interpretable models (green circles) from that data, which approximate the behaviour of the original agents. (c) The interpretable models can replace the original ones (model deployment) or be investigated to better understand the behaviour of the underlying MARL system (model inspection).}
\label{visual_abstract}
\end{figure}

A diagrammatic summary of how MARLeME can be applied to MARL systems is given in Figure~\ref{visual_abstract}, where interpretable models are extracted from MARL system agent data. The interprtable models approximate the behaviour of the MARL agents, and can be further analysed (e.g. through manual inspection, formal verification, or statistical analysis) to provide insight into the behaviour of the underlying MARL system at hand. Moreover, the extracted models can replace the original uninterpretable ones. The three main components of the MARLeME library are described below.

\textbf{TrajectoryHandler}:  
Agents' trajectories are essentially the agents' data MARLeME operates on. In practice, there is a range of ways in which trajectory data may be provided to MARLeME by the user (referred to as `raw data' in Figure~\ref{visual_abstract}), including different data types (e.g. as text files, JSON objects, or in databases), data loading behaviour (e.g. from a server, or locally), and grouping (e.g. incremental, or batch). The \textsc{TrajectoryHandler} module is designed to handle these potential variations, before passing the formatted trajectory data to the \textsc{ModelExtractor} component.

\textbf{ModelExtractor}:
The \textsc{ModelExtractor} component is designed to extract interpretable models from formatted trajectory data provided by the \textsc{TrajectoryHandler}. This component can utilise extraction algorithms for different types of interpretable models, and provide various underlying implementations for these algorithms (e.g. GPU-optimised). The resulting models are represented using \textsc{Team} and \textsc{Agent} components, which are described below. 

\textbf{Team} + \textbf{Agent}: 
Extracted agent models (shown by green circles in Figure~\ref{visual_abstract}) are represented using the \textsc{TeamModel} and \textsc{AgentModel} components. An \textsc{AgentModel} represents a single agent, and a \textsc{TeamModel} represents a group of agents (e.g. teams or sub-teams), giving the user an opportunity to encapsulate agent interactions (e.g. state information sharing). Together, these two classes capture the full spectrum of possible agent interactions: from fully centralised, with all agents represented by a single \textsc{TeamModel} component, to fully decentralised, with all agents running independently (with every agent represented by an \textsc{AgentModel} component). These extracted models can be inspected in order to extract new domain knowledge regarding the corresponding MARL task, or deployed instead their original models, providing more interpretable systems with verifiable properties.

MARLeME can be applied to a vast range of existing MARL systems, by simply logging trajectories of the corresponding MARL agents. Furthermore, MARLeME can be used to extract a wide variety of model types, such as decision trees, or fuzzy rule-based models \cite{Benitez:1997:ANN:2325757.2326154}. This allows successful integration of RL approaches with the well-studied verification, knowledge extraction, and reasoning approaches associated with symbolic models.


\subsection{Abstract Argumentation} \label{abstract_arg}

Extracted models presented in this paper rely on Abstract Argumentation (AA) \citep{dung1995acceptability}, due to the popularity of AA in autonomous agent and multi-agent decision making \citep{gao2016argumentation, ontanon2011argumentation, riveret2019probabilistic}. Argumentation allows reasoning in presence of incomplete and inconsistent knowledge, which is often the case in multi-agent settings, where the agents are unlikely to have complete knowledge of the environment and may have conflicting objectives. In addition, AA-based decision-making models are interpretable and thus highly suitable for interpretable model extraction tasks. Despite these advantages, to the best of our knowledge, this is the first time that AA-based models have been used for interpretable model extraction in RL.

In this work we make use of Value-based Argumentation Frameworks (VAFs) which are based on Abstract Argumentation Frameworks (AFs). Informally, an \(AF=(Arg, Att)\), consists of a set of arguments (\(Arg\)), which are defeasible rules of inference, and a binary attack relation between arguments \(Att \subseteq Arg \times Arg\) that represent conflicts between arguments. In a value-based argumentation framework \(\mathit{VAF} =   (Arg, Att, V, \mathit{val}, \mathit{Valpref}) \),  there are values attached to arguments that represent their relative utility and hence dictate which argument is preferred to another in face of conflict: \( V \) is a set of possible argument values , \( val : Arg \to V \) is a function that assigns values to arguments, and \( \mathit{Valpref} \) is an ordering over these values. Naturally, argumentation frameworks can be represented as directed graphs, where nodes are arguments and an edge from node $A$ to $B$ represent an attack from the former to the latter. 


When representing knowledge using a VAF, it is possible to determine which arguments in the VAF are `winning' or optimal, by defining suitable \textit{semantics} (sets of well-defined rules) that specify whether an argument should be \textit{accepted} (i.e. treated as optimal). The set of accepted arguments of a VAF is referred to as its \textit{extension}. In this work, we make use of the \textit{grounded extension} (GE) semantics when computing acceptable arguments from a VAF, that will be described in the remainder of this section. Details regarding other extension types may be found in \citep{dung1995acceptability}.

Given an AF \( (Arg, Att) \), suppose \( S \subseteq Arg \), then:

\begin{compactitem}
\item{\(S\) attacks an argument \(A\) iff some member of \( S \) attacks \(A\)}
\item{\(S\) is \textit{conflict-free} iff \(S\) attacks none of its members}
\item{\(S\) \textit{defends} an argument \( A \) iff \( S \) attacks all arguments attacking \(A \)}

\item{\(S\) is an \textit{admissible extension} of an AF iff \(S\) is conflict-free and defends all its elements}

\item{\(S \) is a \textit{complete extension} of an AF iff it is an admissible extension and every argument defended by \(S\) belongs to \(S\)}
\item{\(S\) is the \textit{grounded extension} of an AF iff it is the smallest element (with respect to set inclusion) among the complete extensions of the AF.}
\end{compactitem}

Intuitively, a GE represents a `sceptical' solution, consisting only of non-controversial arguments (those that do not attack each other).


\subsection{Abstract Argumentation Agents} \label{aa_agents}

The interpretable agents presented in this work (referred to as \textit{AA-based agents}) rely on VAFs for knowledge representation, and GE semantics for action derivation. The arguments used by these agents are \textit{action arguments}, representing action-based heuristics. An action argument \(A\) has the form `If \textit{condition} holds then \textit{agent should do action}'. In this definition, \textit{condition} is a boolean function of agent state (specifying whether or not argument \(A\) is applicable in that state), \textit{action} specifies the recommended action, and \(agent\) refers to an agent in the team. In this work, all AA-based agents have \(Att\) defined as follows: argument \( A \) attacks argument \( B \) iff they recommend the same action to different agents, or different actions to the same agent. For a given agent, arguments in its argument set \(Arg\) that recommend actions to that agent are referred to as its \textit{primary arguments} in the rest of this work. In our case, \( V \) is the integer set, \( \mathit{Valpref} \) is the standard integer ordering, and \( \mathit{val} \) is a lookup table, mapping arguments to their integer values.

When deriving an action from a state, an AA-based agent computes all arguments in \(Arg\) that are applicable in that state (all arguments for which the argument's condition is true). Then, the agent constructs a VAF from these arguments (using attack construction rules \(Att\), and argument values \(\mathit{val}\)). Finally, the agent uses GE semantics to compute the acceptable primary arguments of the VAF. By definition, all arguments acceptable under GE semantics are non-conflicting, thus, given the agent's definition of \(Att\), all acceptable primary arguments must recommend the same action, which is the action executed by the agent.

Computing applicable arguments, VAF construction, and GE computation are all computationally cheap steps, implying that the AA-based agents successfully scale to complex tasks.

\subsection{Abstract Argumentation Agent Model Extraction}

Extraction of AA-based agents is performed by the algorithm described in Appendix \ref{model_extraction}. This algorithm assumes both the agent argument set \(Arg\) and the attack construction rules \(Att\) are provided by the user, and uses these to derive the argument value ordering \(val\) from the input MARL trajectory data. The agent models are consequently generated using the user-specified \(Arg\) and \(Att\), together with the derived \(\mathit{val}\). The derived value ordering reflects the relative utility of the agent arguments \(Arg\), and can thus serve as an indication of which arguments the agent primarily relies on during action selection, allowing interpretation of agent strategy (as will be shown in Section \ref{strat_analysis}).


\section{Experiments} \label{eval}

We evaluate MARLeME on two well-known RL case studies: Mountain Car \cite{mc_paper} and RoboCup Takeaway \citep{iscen2008new}, using AA-based agents as the extracted models. The remainder of this section describes the two case studies, as well as the corresponding AA-based agent setups.

\subsection{Mountain Car}

Mountain Car was used to show how MARLeME can be applied to simpler tasks, including tasks consisting of a single RL agent. Furthermore, Mountain Car was used to demonstrate that MARLeME can be used in conjunction with other open-source RL tools, such as OpenAI Gym \citep{1606.01540}. The Mountain Car RL task consists of a car that is on a one-dimensional track, positioned between two ``mountains'' \citep{1606.01540}. The goal is to drive up the mountain to the right of the car; however, the car's engine is not strong enough to scale the mountain in a single pass. Therefore, the only way to succeed is to drive back and forth to build up momentum. The RL agent here relies on the Deep Q-Network RL agent implementation~\footnote{https://github.com/adibyte95/Mountain\_car-OpenAI-GYM}, however alternative implementations~\footnote{https://github.com/openai/gym/wiki/Leaderboard} also exists.

Given the simplicity of the environment and the state and action spaces, the argument set \( Arg \) for the Mountain Car AA-based agent was generated using a brute-force approach defined as follows. The Mountain Car state space consists of two variables: \textit{position} and \textit{velocity}, with \( position \in [-1.2, 0.6] \), and \( velocity \in [-0.07, 0.07] \), as well as 3 possible actions: \( actions = \{push\_left, no\_push, push\_right \} \), resembling which direction the car should drive towards. Value ranges for both state variables were split into sets of sub-ranges \( P \) and \( V \), each consisting of 20 equally-sized sub-ranges, with \( P = (p_0, ..., p_{19}) \) and \( V = (v_0, ..., v_{19}) \), where \( p_i \equiv [-1.2 + 0.09 * i, -1.2 + 0.09 * (i+1) ] \), and \( v_i \equiv [-0.07 + 0.007 * i, -0.07 + 0.007 * (i+1) ] \). Using these ranges, the argument set \( Arg \) was defined as \( Arg = \{ \) If \( velocity \in v \) and \( position \in s \) then do \( a \ | \ v \in V, s \in S, a \in actions \} \). Thus, given that \( |P| = 20 \), \( |V| = 20 \), and \( |actions| = 3 \), there were 1200 arguments generated in total.

\subsection{RoboCup Takeaway}

RoboCup Takeaway was used to evaluate MARLeME on a more challenging task, involving a large, continuous state space, multiple interacting agents, and long, variable delays in action effects. RoboCup Takeaway was proposed in \citep{iscen2008new} in order to facilitate RL research in the context of RoboCup Soccer~\footnote{https://rcsoccersim.github.io/}, and focuses on two teams of simulated agents playing the Takeaway game in a two-dimensional virtual soccer stadium. In Takeaway, \( N+1 \) hand-coded keepers are competing with \( N \) independent learning takers on a fixed-size field. Keepers attempt to keep possession of the ball, whereas takers attempt to win possession of the ball. The game consists of a series of episodes, and an episode ends when the ball goes off the field or any taker gets the ball. A new episode starts immediately with all the players reset. We focus here on the 4v3 Takeaway game, consisting of 4 keepers and 3 takers. The MARL taker team consisted of homogeneous, independent learning RL agents relying on the SARSA(\(\lambda\)) algorithm with tile-coding function approximation \citep{sutton1998introduction}. 


When defining arguments for AA-based taker agents, we made use of the work in \citep{aarl_1, aarl_2}, which explored RL agent convergence, and relied on Takeaway as the case study. The mentioned works defined a set of arguments containing useful domain heuristics relevant to the takeaway game, which will be described below.

Every AA-based taker uses the same argument set \( Arg \), consisting of the following arguments (here $i$ is the taker index, ranging from 1 to 3, and $p$ is the keeper index, ranging from 1 to 4): 
\begin{compactitem}
    \item{ \( TackleBall_{i} \) : If \( T_i \) is closest to the ball holder, \(T_i\) should tackle the ball}
    
    \item{ \( OpenKeeper_{i, p} \) : If a keeper \( K_p \) is in a quite `open' position, \( T_i \) should mark this keeper}
    
    \item{\(FarKeeper_{i, p} \): If a keeper \( K_p \) is `far' from all takers, \( T_i \) should mark this keeper}
    
    \item{\(MinAngle_{i, p} \): If the angle between \( T_i \) and a keeper \( K_p \), with vertex at the ball holder, is the smallest, \( T_i \) should mark this keeper}
    
    \item{\( MinDist_{i, p} \) : If \( T_i \) is closest to a keeper  \(K_p \),  \(T_i \) should mark this keeper} 
\end{compactitem}


\section{Results} \label{results}

This section presents the results obtained by evaluating MARLeME on the chosen case studies using the previously described AA-based agent models. Evaluation is both qualitative, demonstrating how we can use MARLeME to inspect the extracted models and extract useful knowledge from them, and quantitative, demonstrating how well extracted models perform on their original tasks, and how closely they approximate their original models.

\subsection{Model Inspection}

Given that extracted models serve as approximations of original models, they may be used to identify and understand the high-level strategies of the original agents, as well as to explain individual action selection. This can be used to extract knowledge about the domain, the nature of agent interactions, and the types of roles agents can take in a team.

In this work, we rely on manual inspection of extracted models. However, MARLeME can be used with many other inspection approaches, depending on the task at hand. In case of larger extracted models that are difficult to analyse manually, inspection may be done by applying automated verification techniques to extracted models, formally proving their properties. Alternatively, extracted models can be analysed empirically, by relying on statistical libraries, such as scikit-learn \cite{scikit-learn}, or Pandas \cite{mckinney-proc-scipy-2010}.  

Our inspection of extracted models presented here is two-fold: firstly (Section \ref{act_select}), we inspect the extracted models at the level of action-selection, showing that extracted models select actions using an interpretable sequence of steps. Secondly (Section \ref{model_vis} and Section \ref{strat_analysis}), we inspect the extracted models at the policy level.

\subsubsection{Action Selection} \label{act_select}

\begin{figure}
\centering
\includegraphics[scale=0.38]{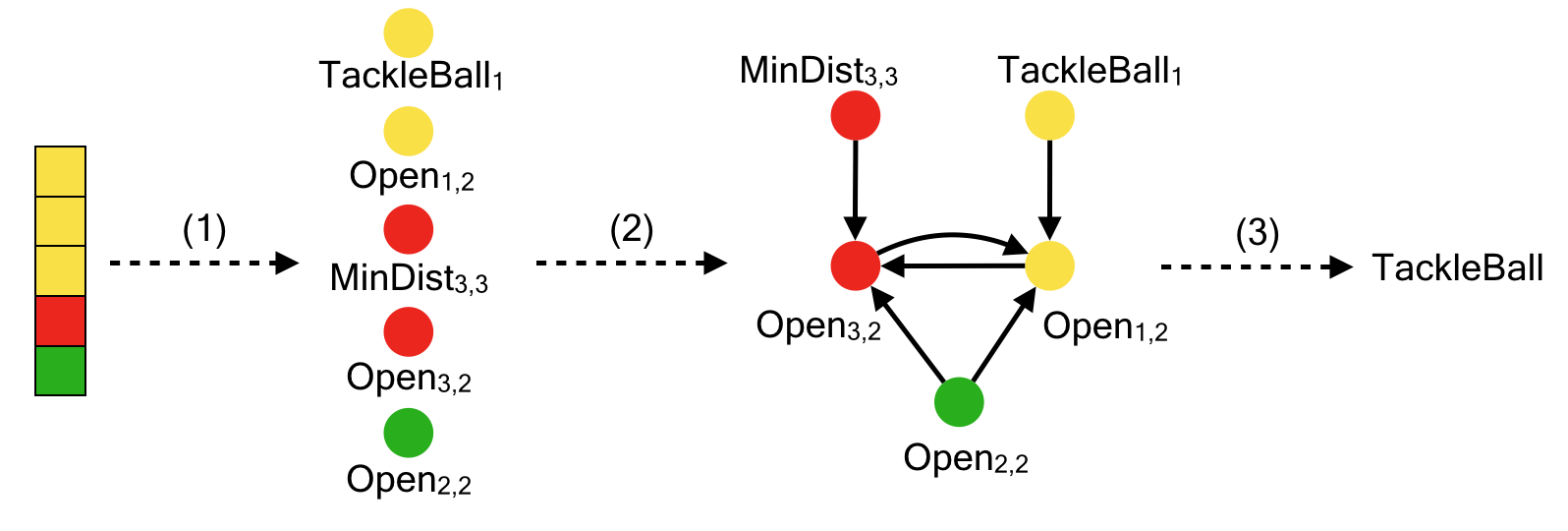} 
\caption{AA-based Taker Agent Model action selection sub-steps for taker $T_1$. (1) Determines the applicable agent arguments for all takers $T_1, T_2, T_3$ (shown by yellow, green, and red circles, respectively), using the input state attributes. (2) Constructs a VAF from the argument set by injecting attacks between arguments (using attack construction rules and the argument value ordering). (3) Derives an action from the VAF for $T_1$, using GE semantics.}
\label{action_select}
\end{figure}

The AA-based agents proposed here make use of an interpretable action selection strategy when deriving an action from a state, compared to their original RL models (an example is shown in Figure \ref{action_select}). Any action selected by an AA-based agent in a given state can be easily traced back to the original set of defined heuristics (\(Arg\)), their interactions (\(Att\)), and their relative values (\(\mathit{val}\)). As described in Section \ref{aa_agents}, an agent's argument set contains arguments recommending actions to the agent itself, as well as arguments recommending actions to its teammates (as shown in Figure \ref{action_select}). Thus, action selection analysis can be used to study the motivations of an individual agent, as well as inter-agent interactions, that influence action selection.


\subsubsection{Model Visualisation} \label{model_vis}

\begin{figure}
\centering     
\subfigure[Extracted Model Policy]{\label{extr_policy}\includegraphics[width=60mm]{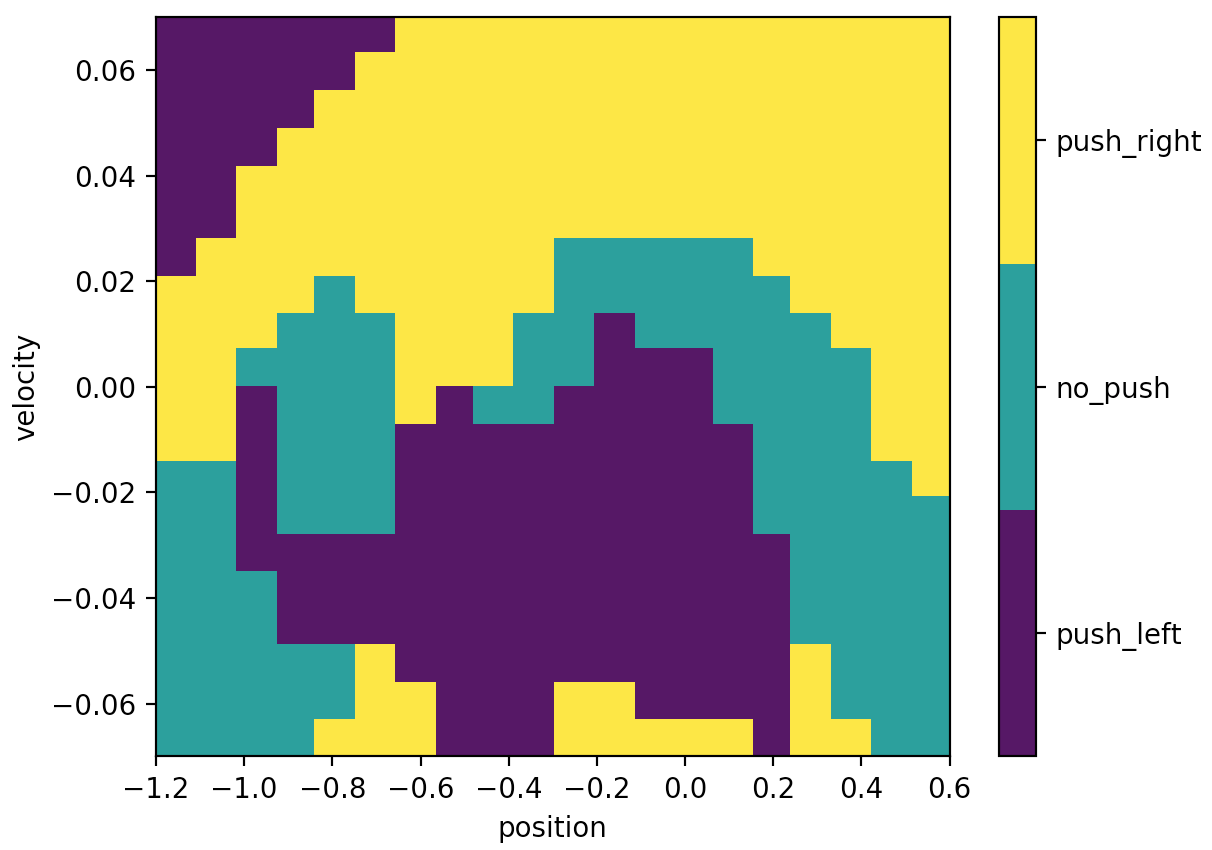}}
\subfigure[Original Model Policy]{\label{orig_policy}\includegraphics[width=60mm]{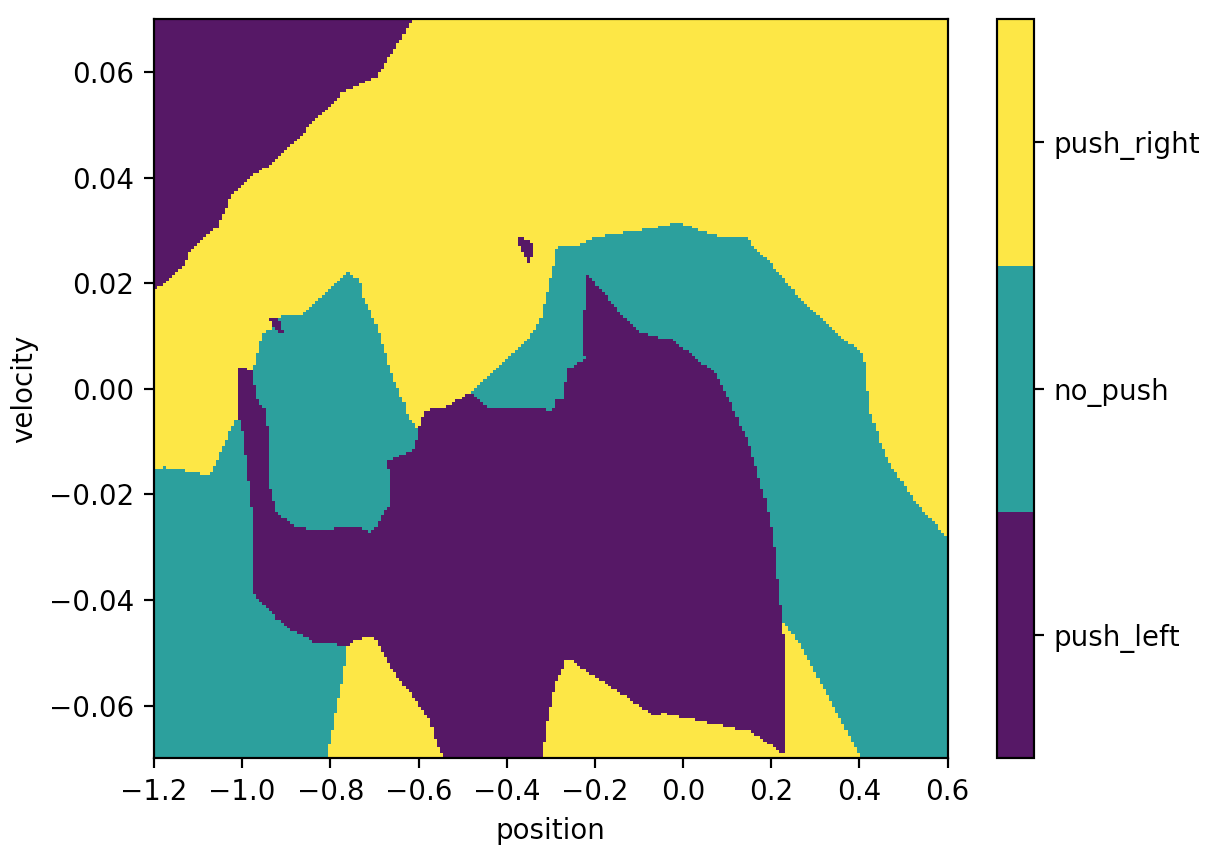}}
\caption{Policy visualisations of the extracted and original Mountain Car models, showing policy outputs in all possible environment states.}
\end{figure}

In case of relatively simple tasks, such as Mountain Car, it is possible to visualise the action-selection policy of the extracted model directly, by plotting policy actions in all possible states, as shown in Figure \ref{extr_policy}. This visualisation allows direct exploration of the extracted model, allowing users to derive useful information by observing model behaviour in various states. 

Furthermore, visualisation allows directly comparing the extracted model policy to the original model policy (shown in Figure \ref{orig_policy}), allowing a user to easily determine where and how the extracted model differs from the original one.

\subsubsection{Strategy Analysis} \label{strat_analysis}

In case of more challenging tasks, such as RoboCup Takeaway, the extracted models may be inspected by studying their constituent components. AA-based agent models offer an intuitive way of analysing agent strategies as a whole, exploring individual agent behaviour, and cooperative team behaviour, by using their argument value orderings \(val\). As described in Section \ref{aa_agents}, an agent chooses an action recommended by its primary arguments. Given that arguments with higher values defeat arguments with lower values, high-valued primary arguments represent the main heuristics used by an agent during decision-making. Thus, analysing the highest-valued primary arguments can be used to explore individual agent strategies, and their behaviour. Table \ref{avo} gives the top 5 highest-valued primary arguments for every taker.

\begin{table}
\centering
\begin{tabular}{| c | c | c |} \hline

\textbf{Agent 1} & \textbf{Agent 2} & \textbf{Agent 3}  \\ \hline

$TackleBall_1$	&	$TackleBall_2$ 	& 	$TackleBall_3$ 		\\ \hline
$MinAngle_{1, 3} $	&	$MinAngle_{2, 3} $	& 	$MinAngle_{3, 4} $			\\ \hline
$MinAngle_{1, 2} $	&	$MinDist_{2, 3} $	& 	$MinDist_{3, 4}	$		\\ \hline
$MinDist_{1, 2}	$	&	$OpenKeeper_{2, 3}$	& 	$OpenKeeper_{3, 4}$			\\ \hline
$MinDist_{1, 3}	$	&	$FarKeeper_{2, 3}$	&	$FarKeeper_{3, 4}$	\\ \hline
\end{tabular}
\caption{Highest-valued primary arguments of the AA-based Taker agents}
\label{avo}
\end{table}

For all three agents, their \textit{TackleBall} argument has the highest value, implying that tackling the ball when closest to the keeper is of paramount importance for every agent. For Agent 1, the four remaining arguments show that this agent focuses on tackling \( K_2 \), or \( K_3 \), if it is `closest' to them (closest by angle, or distance). For Agents 2 and 3, their four remaining arguments all recommend marking \( K_3 \) and \( K_4 \) (respectively), implying that these agents focus on tackling these keepers throughout the game, preventing the ball-holder from passing to them.

The above analysis demonstrates the specialised roles taken by the different agents (e.g. different agents in the above example focus on tackling different keepers). As discussed previously, this knowledge may be used to study the Takeaway game in more detail, and attempt to synthesise winning agent strategies. For instance, the above example demonstrates that a possible strategy is for one taker to tackle the ball-holder, and the other two to `spread out' and tackle distant keepers, in order to prevent them from receiving the ball.

\subsection{Model Deployment}

The utility of extracted models was also evaluated quantitatively, by replacing the original RL agents with their corresponding AA-based agents, and comparing the task performance of the two approaches. Task performance was measured by recording average episode duration for 1000 episodes. In both case studies, a shorter episode duration signifies better performance.  The above setup was run on a MacBook Pro computer with a 4-core 2.5GHz Intel Core i7 processor, and 16GB of main memory. The performances of the different approaches are shown in Table \ref{deployment}. 

\begin{table}
\centering
\begin{tabular}{| c | c | c |}
\hline
& \textbf{Original Models} & \textbf{Extracted Models} \\ \hline

\textbf{Mountain Car} &  55.3 +/- 11.3 ms & 2.96 +/- 0.4 ms \\ \hline

\textbf{Takeaway} &  9.55 +/- 3.3 s & 10.71 +/- 3.8 s \\ \hline

\end{tabular}
\caption{Task performance of the original models, and the extracted models}
\label{deployment}
\end{table}

Increased interpretability of extracted models typically comes at a cost of their reduced flexibility, implying that they may perform sub-optimal decision making, compared to the original models, and therefore incur a reduction in performance, as is the case with AA-based taker agents. 

On the other hand, extracted models are often simpler, and therefore more lightweight, compared to the original models. This implies that extracted models often take less time to compute actions from states. In case of simpler tasks such as Mountain Car, where the reduction in flexibility is not substantial, this speedup may have a large effect on task performance (a factor of 18 speedup in case of Mountain Car). 

Overall, extracted models offer a substantial increase in interpretability, at the expense of reduced flexibility, which may percolate into reduced performance when considering more challenging tasks.

\subsection{Model Fidelity}

Finally, we evaluate fidelity of extracted models using the \textit{0-1 loss} \citep{ross2011reduction} computed from 1000 episodes of trajectory data, in order to assess the degree to which we can rely on the extracted models when studying the original ones. In case of Mountain Car, the extracted model achieved a high fidelity score of 0.93. In case of the more challenging Takeaway task, agent 1 and agent 3 achieved high fidelity scores of 0.86 and 0.84 (respectively), whilst agent 2 achieved a relatively lower fidelity score of 0.68. These results indicate that model extraction for agent 2 could be further improved by using a more flexible extracted model (e.g. an AA-based agent model with a larger argument set \(Arg\)), consequently making extracted model interpretation even more meaningful. 



\section{Conclusions} \label{conclusion}

We introduce MARLeME, a Multi-Agent Reinforcement Learning Model Extraction library, designed to improve interpretability of MARL systems by approximating MARL agents with symbolic models, which can be used to study the behaviour of the underlying agents, as well as to replace all/some of them. MARLeME can be applied to a vast range of existing MARL systems, and can be used with a wide variety of symbolic model types. Thus, MARLeME allows successful integration of RL approaches with the well-studied verification, knowledge extraction, and reasoning approaches associated with symbolic models. 

In this work, we focus on applying MARLeME to trained MARL agents, in order to extract useful domain knowledge by studying their behaviour. That said, given its ease-of-use, flexibility, and extendibility, MARLeME can be applied in a broader range of RL scenarios, depending on the task at hand. For instance, MARLeME can be used to periodically inspect RL models over time during agent training, seeing how MARL agent policies evolve, and exploring their convergence. 

With the rapidly increasing interest in MARL systems and the development of associated tools (such as the recently released OpenAI MARL environments repository ~\footnote{https://openai.com/blog/emergent-tool-use/}), we believe MARLeME can play a fundamental role in enriching such MARL systems and tools with explainability and interpretability.


\clearpage


\bibliographystyle{plainnat}
\bibliography{references}

\clearpage

\appendix

\section{AA-based Model Extraction} \label{model_extraction} 

The algorithm introduced in this work for extracting AA-based agent models from MARL agent trajectories is given in Algorithm \ref{getOrdering}. Algorithm \ref{getOrdering} assumes that the set of candidate arguments (\textit{Arguments}) is fixed, and is passed as an input parameter. The algorithm focuses on deriving an argument ordering (\textit{extractedOrdering}) using the input trajectories (\textit{Trajectories}) and a default input ordering (\textit{DefaultOrdering}).  

\begin{algorithm} 
\caption{extractOrdering($Trajectories, Arguments, DefaultOrdering$)} \label{getOrdering}
\begin{algorithmic}[1]

\State $preferenceGraph = newArgPreferenceGraph()$

\For{$(state, action) \in Trajectories$}
	\State $applicableArguments = getApplicableArguments(Arguments, state)$
	\State $ relevantArguments = \{ arg \in applicableArgs \ | \ arg.Conclusion == action \} $
	\State $irrelevantArguments = applicableArguments - relevantArguments$
	\For{ $ relevantArg \in relevantArguments$ }
		\For{$irrelevantArg \in irrelevantArguments$}
			\State $preferenceGraph.incrementEdge(relevantArg, irrelevantArg)$
		\EndFor
	\EndFor
\EndFor

\State $DAG = convertToAcyclic(preferenceGraph)$
\State $extractedOrdering = topologicalSort(DAG.nodes, DAG.edges, DefaultOrdering)$
\State \textbf{return} $extractedOrdering$
\end{algorithmic}
\end{algorithm}

\begin{figure}
\centering
\includegraphics[scale=0.35]{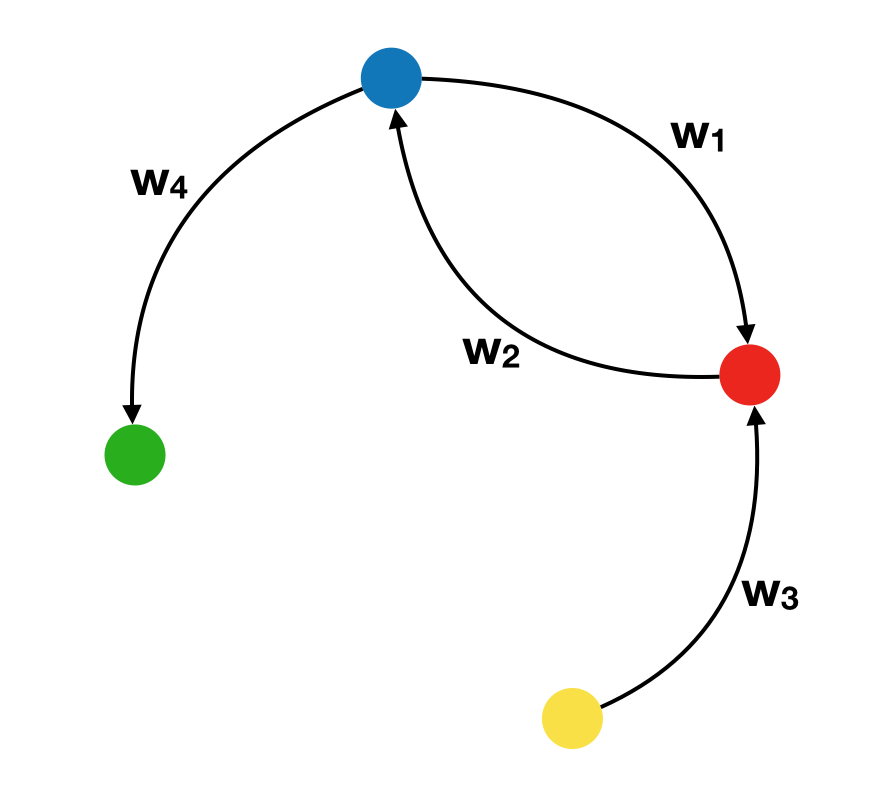}
\caption{Simple example of an APG consisting of four arguments (represented by coloured circles), and weighted preference edges with weights $w_1, ..., w_4$. }
\label{argPref}
\end{figure}

Algorithm \ref{getOrdering} relies on topological sorting and operates on a novel Argument Preference Graph (APG) structure (also introduced in this work). The fundamental idea behind the algorithm is that for a given pair of arguments \( (A,B) \) and a RL agent trajectory, the argument that is `in closer agreement' with the agent contains relatively more useful information, and should thus have a relatively higher value. These pair-wise argument preferences are stored in an APG. An APG is a weighted directed graph (with non-negative weights) in which the nodes represent arguments, and weighted edges represent preferences between arguments. Figure \ref{argPref} shows an example of an APG constructed by the \textit{extractOrdering} method.

The relative argument utility is computed by iterating over all \( (state, action) \) pairs of input agent trajectories. For every such \( (state, action) \) pair, the APG is updated by incrementing the weights of directed edges between all pairs of arguments \( (A,B) \), where both \( A \) and \( B \) are applicable in \( state \), \( A \) was in agreement with the agent (recommended the same action), and \(B\) was not (recommended a different action). Thus, a directed edge with weight \(w\) from argument \( A \) to argument \(B\) in an APG implies that \(A\) was applicable and in agreement with the agent, whilst argument \(B\) was applicable was not in agreement with the agent, for \(w\) different trajectory states. A high value of weight \(w\) of a directed edge \( (A,B) \) thus signifies that argument \( A \) frequently suggested more relevant information than argument \( B \), implying that argument \( A \) should have a higher value.

Once the APG is constructed, an argument ordering is extracted from it by first converting it into a Directed Acyclic Graph (DAG) (by calling the \textit{convertToAcyclic} method), and then topologically sorting the DAG (by calling the \textit{topologicalSort} method). Graph cycle removal is achieved by relying on a pruning heuristic, which removes all edges of weight less than a given pruning value \(p\) from the graph. Topological sorting is achieved using a slight variation of Kahn's algorithm \citep{Kahn:1962:TSL:368996.369025}, where the node extraction order relies on the default argument ordering given by the user, instead of relying on stack or queue structures.

Overall, the ordering extraction algorithm relies on both knowledge derived from the trajectory data (in the form of an APG), and on heuristics injected by the user (the default ordering). The utilised pruning approach is advantageous since it provides a straightforward way of controlling the tradeoff between the importance of prior knowledge and extracted knowledge, by using the pruning parameter \(p\). A higher \(p\) value removes more edges and yields a sparser APG with more possible argument orderings, leading to greater reliance on the default ordering. A lower \(p\) value has the reverse effect.






\end{document}